%% file: EquiFlow.tex
\documentclass[letterpaper]{article} % DO NOT CHANGE THIS
\usepackage{EquiFlow}  % DO NOT CHANGE THIS [final]
\usepackage{times}  % DO NOT CHANGE THIS
\usepackage{helvet}  % DO NOT CHANGE THIS
\usepackage{courier}  % DO NOT CHANGE THIS
\usepackage[hyphens]{url}  % DO NOT CHANGE THIS
\usepackage{graphicx} % DO NOT CHANGE THIS
\usepackage{natbib}  % DO NOT CHANGE THIS AND DO NOT ADD ANY OPTIONS TO IT
\usepackage{caption} % DO NOT CHANGE THIS AND DO NOT ADD ANY OPTIONS TO IT
\usepackage[symbol]{footmisc}
\usepackage{amsfonts}
\usepackage{amsmath}
\usepackage{algorithm}
\usepackage{algorithmic}
\usepackage{booktabs}
\usepackage{multirow}
\usepackage{appendix}

\usepackage{float}
\usepackage{newfloat}
\usepackage{listings}
\DeclareCaptionStyle{ruled}{labelfont=normalfont,labelsep=colon,strut=off} % DO NOT CHANGE THIS
\floatstyle{ruled}
\newfloat{listing}{tb}{lst}{}
\floatname{listing}{Listing}

\title{EquiFlow: Equivariant Conditional Flow Matching with Optimal Transport for 3D Molecular Conformation Prediction}

\author {
    Qingwen Tian\textsuperscript{\rm 1,2}\equalcontrib,
    Yuxin Xu\textsuperscript{\rm 2}\equalcontrib,
    Yixuan Yang\textsuperscript{\rm 2},
    Zhen Wang\textsuperscript{\rm 2},
    Ziqi Liu\textsuperscript{\rm 2,3},\\
    Pengju Yan\textsuperscript{\rm 2}\corrauthor,
    Xiaolin Li\textsuperscript{\rm 2}\corrauthor
}
\affiliations {
    \textsuperscript{\rm 1}Zhejiang University of Technology\\
    \textsuperscript{\rm 2}HIM, Chinese Academy of Sciences\\
    \textsuperscript{\rm 3}University of Chinese Academy of Sciences\\
    tqw@zjut.edu.cn, 
    xux@live.ca, 
    yangyixuan@him.cas.cn, 
    wangzhen@him.cas.cn, 
    liuziqi22@mails.ucas.ac.cn, 
    yanpengju@gmail.com, 
    xiaolinli@ieee.org
}

\usepackage{bibentry}
\begin{document}
\maketitle

\begin{abstract}
% Molecular 3D conformations determine their interactions with other molecules or protein surfaces, playing a crucial role in AI-assisted drug discovery. 
Molecular 3D conformations play a key role in determining how molecules interact with other molecules or protein surfaces. Recent deep learning advancements have improved conformation prediction, but slow training speeds and difficulties in utilizing high-degree features limit performance. We propose EquiFlow, an equivariant conditional flow matching model with optimal transport. EquiFlow uniquely applies conditional flow matching in molecular 3D conformation prediction, leveraging simulation-free training to address slow training speeds. It uses a modified Equiformer model to encode Cartesian molecular conformations along with their atomic and bond properties into higher-degree embeddings. Additionally, EquiFlow employs an ODE solver, providing faster inference speeds compared to diffusion models with SDEs. Experiments on the QM9 dataset show that EquiFlow predicts small molecule conformations more accurately than current state-of-the-art models.
\end{abstract}

\section{Introduction}
3D molecular conformation prediction involves using computational methods and deep learning algorithms to determine the 3D spatial arrangement of molecules. In this context, atoms are represented by their Cartesian coordinates, and the 3D conformations of a molecule are critical for understanding its functionality, reactivity, and physical properties. The 3D conformations of a molecule dictate how it interacts with other molecules or protein surfaces, thereby influencing its chemical activity and function. This prediction has broad applications in computational drug and material design \cite{thomas2018tensor, gebauer2022inverse, jing2020learning, batzner2022}.

Traditional methods for predicting molecular conformations, such as those based on classical force fields \cite{wang2004development} or quantum mechanical calculations, are accurate but computationally intensive and difficult to apply to large datasets. Additionally, methods like molecular dynamics or Markov Chain Monte Carlo are prohibitively expensive for large molecules, often requiring extensive runtime to produce reliable results and may not resolve molecular symmetry issues.

Recently, deep learning has achieved significant advancements in fields such as image recognition and natural language processing. Consequently, more researchers are now exploring its application in molecular conformation prediction. The success of deep learning is largely attributed to its ability to employ structural symmetry. For instance, convolutional neural networks (CNNs) can process 2D images and recognize patterns efficiently irrespective of their positions, highlighting the importance of translational equivariance. Similarly, in processing 3D atomic graphs for molecules, the relevant inductive bias pertains to the 3D Euclidean group E(3), which encompasses equivariance under operations involving translation, rotation, and reflection \cite{liao2023equiformer}. As a result, certain properties of atomic systems should remain invariant under specific transformations. For example, when a system rotates, associated physical quantities like forces must rotate accordingly. To address this requirement, neural networks with invariance and equivariance properties have been developed. These networks have proven to be efficient tools for handling complex atomic graph data by effectively integrating these geometric principles.

In invariant neural networks, researchers aim to extract information such as distances and angles from 3D graph data that remain unchanged under rotations and translations, thereby enhancing the representation capability of the graphs \cite{simm2020generative, xu2021end, shi2021learning, ganea2021geomol}. These methods avoid directly modeling complex atomic coordinates by exploiting intermediate geometric variables that are invariant to rotations and translations. However, reliance on these intermediate variables may restrict the model's flexibility and limit its applicability during both training and inference. Therefore, directly modeling atomic coordinates while maintaining translational and rotational invariance becomes an ideal solution \cite{xu2022geodiff}, necessitating the use of equivariant neural networks. In the task of 3D molecular modeling, researchers have achieved significant results by constructing equivariant graph neural networks to predict or generate 3D molecular conformations. For instance, inspired by the diffusion process in classical non-equilibrium thermodynamics, GeoDiff \cite{xu2022geodiff} models each atom as a particle and learns to simulate the diffusion process (from noise distribution to stable conformations) as a Markov chain to predict 3D molecular conformations. However, diffusion models (DMs) require substantial computational resources during both training and inference. Torsional Diffusion \cite{jing2022torsional} addresses this issue by predicting the hypertorus on the torsion angles of each 3D molecular conformation instead of directly predicting their atomic coordinates. This method, however, relies on RDkit to generate rigid substructures for each molecule, hence it is not an end-to-end model. Inspired by recently proposed flow matching models \cite{lipman2023flow}, EquiFM \cite{song2024equivariant} employs a novel flow matching training objective for geometric generation modeling in molecular generation. It utilizes Equivariant Graph Neural Networks (EGNN) \cite{satorras2021n} to parameterize the vector field, trained via flow matching to satisfy translational and rotational equivariance constraints. However, these methods often rely on constructing equivariant graph neural networks based on low-degree features, limiting the benefit from information-rich high-degree features.

To avoid these limitations, we propose EquiFlow, an equivariant conditional flow matching (CFM) model with optimal transport (OT) for 3D molecular conformation prediction. EquiFlow is the first method to apply CFM to molecular conformation prediction, leveraging the efficient training and inference capabilities of CFM. It integrates a modified Equiformer model to encode Cartesian molecular conformations, fully adopting atomic type features and bond features, and facilitating effective interaction with higher-degree molecular features. Moreover, EquiFlow leverages an ODE solver, thereby facilitating more rapid inference speeds in comparison to DMs that utilize SDE. The main contributions of this work are as follows:

\begin{itemize}
    \item Training the model with an OT flow objective that directly predicts the vector field around the atomic coordinates of molecules, resulting in stable and efficient training, high accuracy, and excellent diversity.
    \item Encoding Cartesian molecular conformations with modified Equiformer, leveraging atomic type and bond features, enabling effective interaction with higher-degree molecular features, and ensuring translational and rotational equivariance.
    % \item Extensive experiments demonstrate that EquiFlow outperforms existing advanced machine learning methods in molecular conformation prediction, achieving state-of-the-art performance on the single-conformation QM9 and multi-conformation GEOM-QM9 datasets.
    \item Introducing the use of OT-CFM for the first time in the task of 3D molecular conformation prediction, providing a novel approach that combines optimal transport theory with flow matching to significantly improve the performance of conformation prediction models.
\end{itemize}

\begin{figure*}[h]
    \centering
    \includegraphics[width=\textwidth]{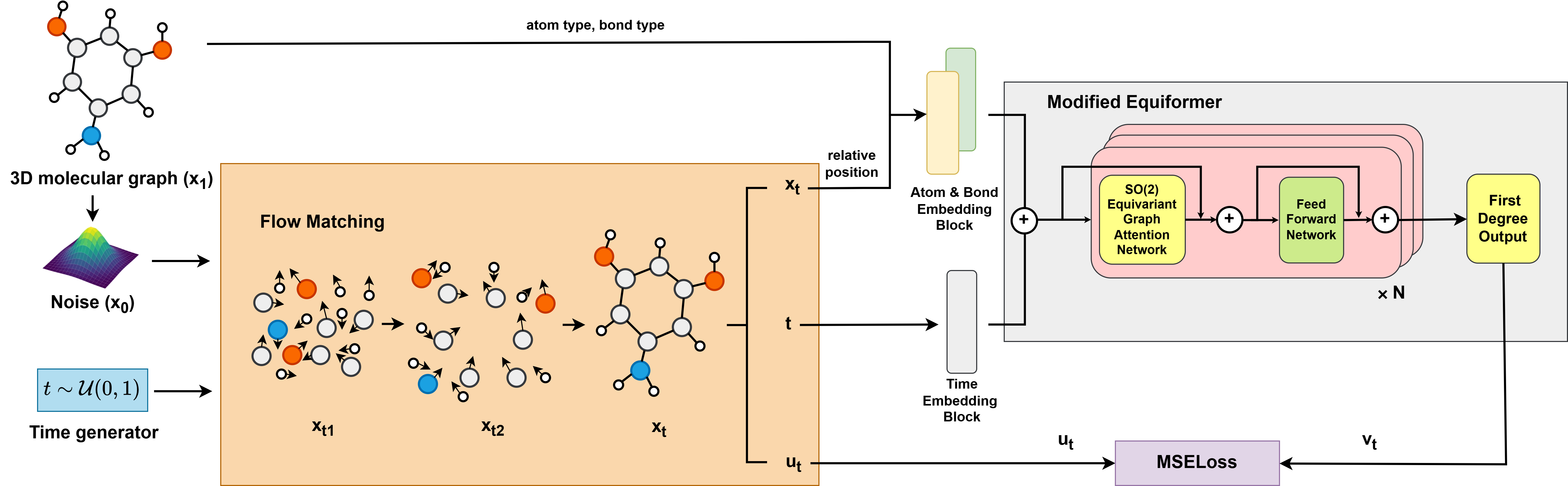}
    % \caption{The architecture of ConfromFlow. The architecture involves using fixed atom and bond types as conditions, followed by Flow Matching on the atomic coordinates to produce \(x_T\), \(T\), and \(u_T\). After embedding the relative position features and time features \(T\) along with the conditions, a modified Equiformer is employed to predict the vector field \(v_T\) around the atoms. Using Mean Squared Error (MSE) loss, the predicted vector field \(v_T\) is then fitted to the ground truth vector field \(u_T\) obtained from Flow Matching. For training procedure, see Algorithm \ref{alg:train}, for sampling procedure, see Algorithm \ref{alg:sample}.}
    \caption{The architecture of ConfromFlow. The architecture involves using fixed atom and bond types as conditions, followed by Flow Matching on the atomic coordinates to produce \(x_t\), \(t\), and \(u_t\). After embedding the relative position features and time features along with the conditions, a modified Equiformer is employed to predict the vector field \(v_t\) around the atoms. Using Mean Squared Error (MSE) loss, the predicted vector field \(v_T\) is then fitted to the ground truth vector field \(u_T\) obtained from Flow Matching. For training procedure, see Algorithm \ref{alg:train}, for sampling procedure, see Algorithm \ref{alg:sample}.}
    \label{fig:Conformation}
\end{figure*}

\section{Related Works}
\subsection{Diffusion Model and Conditional Flow Matching}
DMs \cite{sohl2015deep, song2019generative, ho2020denoising} have recently achieved success in high-dimensional statistics \cite{rombach2022high}, language modeling \cite{li2022diffusion}, and equivariant representations \cite{hoogeboom2022equivariant}. In the 3D molecular field, GeoDiff \cite{xu2022geodiff} and EDM \cite{hoogeboom2022equivariant} adopt DMs to learn score-based models for predicting molecular conformations. However, these approaches often face challenges such as unstable probability dynamics and inefficient sampling, which hinder their overall effectiveness. Torsional Diffusion \cite{jing2022torsional} addresses these issues by decomposing molecules into rigid substructures, thereby reducing the computational burden to calculating only the torsion angles that connect these substructures. Despite its advantages, this method requires the pre-computation of rigid substructures, which can be a limitation for certain applications.

Normalizing flows (NFs) \cite{tabak2010density, tabak2013family, rezende2015variational, papamakarios2021normalizing} are powerful generative neural networks that construct a reversible and efficient differentiable mapping \cite{tong2024improving} between a fixed distribution (e.g., standard normal) and the data distribution \cite{rezende2015variational} through exact likelihood estimation. Traditional NFs design this mapping as a static composition of reversible modules, while continuous normalizing flows (CNFs) represent this mapping through neural ODEs \cite{chen2018neural}. However, CNFs face challenges in training and handling large datasets \cite{chen2018neural, grathwohl2018scalable, onken2021ot}. Flow Matching (FM) \cite{lipman2023flow} generates high-quality samples and stabilizes CNFs training, but its reliance on the Gaussian source distribution makes it less convenient for broader applications. CFM \cite{tong2024improving} addresses this problem by extending existing diffusion and FM methods. For example, EquiFM \cite{song2024equivariant} and MolFlow \cite{irwin2024molflow} use CFM for 3D molecular generation, achieving superior performance across multiple molecular generation benchmarks and improving sampling speed.

In this paper, we apply the CFM method to molecular conformation prediction, directly predicting the vector field around molecular atomic coordinates and achieving stable and efficient training.

\subsection{Invariant and Equivariant Neural Networks}
Invariant neural networks model molecules using intermediate geometric variables, including interatomic distances \cite{simm2020generative, xu2021end, shi2021learning} and bond features \cite{ganea2021geomol}, which are invariant to translation and rotation. These methods avoid directly modeling complex atomic coordinates, but their reliance on intermediate geometric variables limits their flexibility during training or inference.

Equivariant neural networks \cite{thomas2018tensor, thölke2022equivariant, le2022equivariant, liao2023equiformer, liao2024equiformerv2} operate on type-L vectors and geometric tensors. These models utilize tensor field networks (TFN) \cite{thomas2018tensor}, which employ spherical harmonics and irreducible representations to ensure three-dimensional translational and rotational equivariance. Methods based on equivariant transformers \cite{thölke2022equivariant, le2022equivariant} used dot product attention \cite{vaswani2017attention} and linear message passing, and adopted a specialized architecture that considered only type-0 and type-1 vectors. Equiformer \cite{liao2023equiformer} integrates MLP attention, non-linear messaging \cite{gilmer2017neural, sanchez2020learning}, and higher-degree vectors to incorporate higher-degree tensor product interactions, achieving enhanced performance. EquiformerV2 \cite{liao2024equiformerv2} builds on Equiformer, replacing SO(3) convolutions with eSCN convolutions and optimizing the model architecture to enhance performance in higher-degree contexts.

We enhance the embedding blocks of EquiformerV2 \cite{liao2024equiformerv2} to encode Cartesian molecular conformations while ensuring both translational and rotational invariance. This improvement efficiently captures feature information and facilitates effective interaction with higher-degree molecular features.

\subsection{3D Molecular conformation Prediction}
Cheminformatics and deep learning are the two most commonly used approaches for 3D molecular conformation prediction. The former leverages chemical heuristics, rules, and databases to achieve significantly faster generation. However, while these methods efficiently model constrained degrees of freedom, they fall short in capturing the full energy landscape. Notable examples include the commercial software OMEGA \cite{hawkins2010omega} and the open-source RDKit ETKDG \cite{landrum2013rdkit}. On the other hand, deep learning methods exploit molecular structural symmetry to introduce inductive biases, enhancing their predictive capabilities. Examples include GeoMol \cite{ganea2021geomol}, GeoDiff \cite{xu2022geodiff}, and Torsional Diffusion \cite{jing2022torsional}.

EquiFlow stands out by effectively simulating the complex distribution of molecular conformations and accurately predicting the 3D structure of molecules.

\section{Methodology}
In this section, we first define the problem of predicting the 3D conformation of molecules. Subsequently, we detail the components of EquiFlow, including an equivariant CFM with OT and a modified equiformer. The framework of EquiFlow is illustrated in Figure \ref{fig:Conformation}.

\subsection{Problem Definition}
A molecule can be represented as an undirected graph \( {G=⟨X, E⟩} \), where \( {X=(X_1, X_2,...,X_k)\in\mathbb{R}^{k\times3}} \) denotes the coordinate matrix of \( k \) atoms, and \( {E=(e_{ij},(i,j)\in|X|\times|X|)} \) denotes the feature matrix of the edges between atoms. Our goal is to model the known atomic coordinates of a molecule while keeping the atomic types fixed, in order to accurately predict the 3D conformation of the molecule, encompassing both diversity and accuracy.

\subsection{Equivariant Conditional Flow Matching with Optimal Transport}

\subsubsection{Normalizing Flows (NFs).}
Let $ x_0 $ denotes data points from a prior distribution $ p_0 $, and $ x_1 $ denotes samples from an output distribution $ p_1 $. NFs \cite{rezende2015variational, papamakarios2021normalizing} provide a powerful framework for learning complex probability distributions through the concept of invertible transformations ($ f_\theta: \mathbb{R}^{k\times3} \rightarrow \mathbb{R}^{k\times3} $). It transforms samples from $ p_0(x_0) $ to $ p_1(x_1) $, known as the forward distribution, and utilizes the change of variables formula:
\begin{equation}
p_1(x_1) = p_0(f^{-1}(x_1)) \left| \det \left( \frac{\partial f_\theta^{-1}(x_1)}{\partial x_1} \right) \right|
\end{equation}
where $ \frac{\partial f_\theta^{-1}(x_1)}{\partial x_1} $ represents the Jacobian matrix of the inverse transformation $ f_\theta^{-1} $. In essence, NFs enable the continuous reshaping of a simpler, known distribution into a more intricate, unknown distribution, while preserving invertibility and allowing for exact likelihood evaluation.

\subsubsection{Continuous Normalizing Flows (CNFs).}
First, we define some notations: the time-dependent probability density path $ p_{t \in [0,1]}: \mathbb{R}^{k\times3} \rightarrow \mathbb{R}_{>0}$, the time-dependent vector field $ v_{t \in [0,1]}: \mathbb{R}^{k\times3} \rightarrow \mathbb{R}^{k\times3} $, and the time-dependent flow $ f_\theta(x,t): \mathbb{R}^{k\times3} \times [0,1] \rightarrow \mathbb{R}^{k\times3} $. CNFs \cite{chen2018neural} are a class of NFs defined by an ODE:
\begin{equation}
df_\theta(x,t) = v_\theta(f_\theta(x,t), t) \, dt
\end{equation}
with the initial condition:
\begin{equation}
f_\theta(x,0) = x
\end{equation}
where the time-dependent vector field uniquely determines the time-dependent flow. \cite{chen2018neural} proposed using an ODE solver to train CNFs. However, due to the need for numerical ODE simulation, CNFs are challenging to train.

\subsubsection{Optimal Transport Between Conformations.}

\begin{figure}[h]
    \centering
    \includegraphics[width=1.0\linewidth]{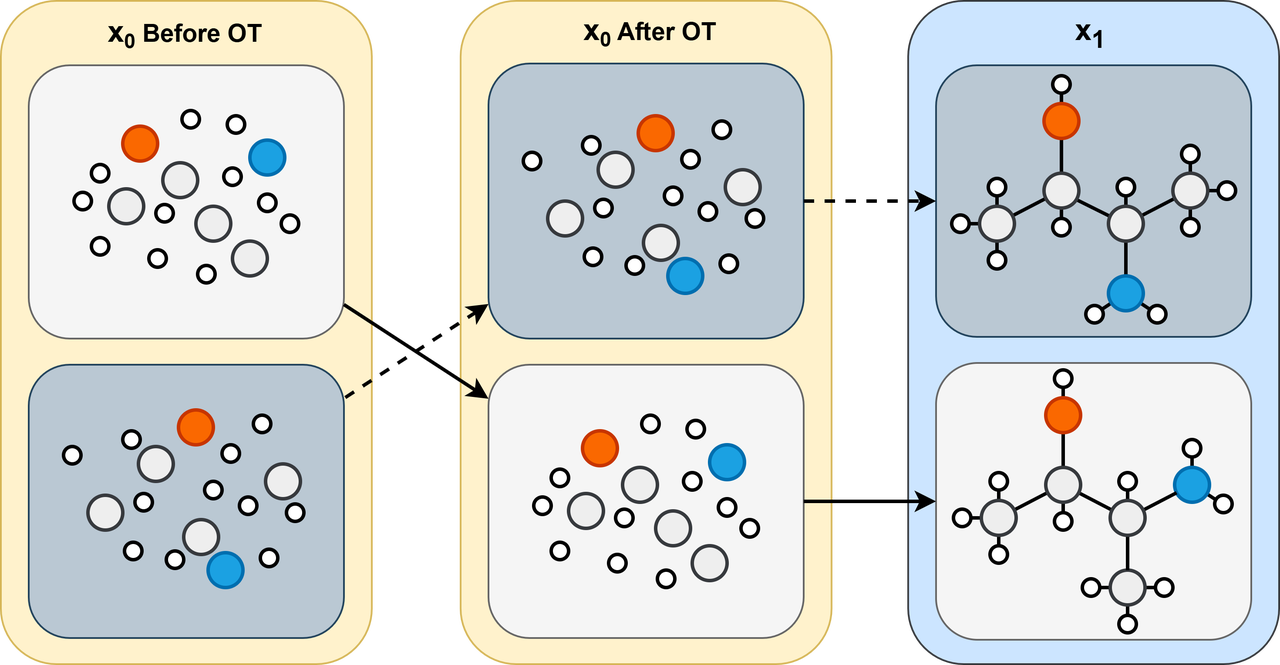}
    \caption{The process of Equivariant OT-CFM. We perform OT between different conformations of the same molecule to obtain the mapping that minimizes the transport cost between the Gaussian noise coordinates \(x_0\) and the true conformation coordinates \(x_1\). Following this, we calculate the conditional probability path and the corresponding conditional vector field during the CFM process. Note that both \(x_0\) and \(x_1\) need to be centered using the Zero Center-of-Mass (Zero CoM) operation to ensure translational equivariance, and the Kabsch algorithm \cite{kabsch1976solution} is used for rotational alignment.}
    \label{fig:ot}
\end{figure}

\cite{lipman2023flow} proposed the Flow Matching objective to train CNFs without simulation, regressing the neural network $ v_\theta(x,t) $ to a target vector field $ u_t(x) $:
\begin{equation}
\mathcal{L}_{\mathrm{FM}}(\theta) = \mathbb{E}_{t, p_{t}(x)} \left\| v_\theta(x, t) - u_t(x) \right\|^2
\end{equation}

However, this objective requires prior knowledge of the vector field $ u_t(x) $ and the corresponding probability density path $ p_t(x) $, which is difficult to achieve.

To address the computational challenges of Flow Matching objectives, it has been shown by \cite{lipman2023flow} that utilizing CFM loss yields gradients equivalent to traditional methods. Moreover, \cite{tong2024improving} has successfully applied CFM loss to achieve effective results (see details in Appendix B):
\begin{equation}
\mathcal{L}_{\mathrm{CFM}}(\theta) = \mathbb{E}_{t, p_{t}(x|z)} \left\|v_{\theta}(x, t) - u_{t}(x|z)\right\|^2
\end{equation}
where the conditional distribution is defined by the OT map $ \pi(x_0, x_1) $, which is determined by the 2-Wasserstein distance:
\begin{equation}
W_{2}^{2} = \inf_{\pi \in \Pi} \int c(x_{0}, x_{1}) \, \pi(dx_{0}, dx_{1})
\end{equation}
here, $ \Pi $ represents the set of all couplings, and $ c(x_0,x_1) = \|x_0 - x_1\|^2 $ denotes the squared Euclidean distance used as the cost function to obtain the OT map matrix $ M $.

In this work, we focus on predicting atomic-level features, emphasizing fixed atomic type characteristics while concentrating on the OT mapping (Figure \ref{fig:ot}) of molecular coordinates $x$. As seen in Appendix B, previous studies \cite{lipman2023flow, tong2024improving} calculated the squared Euclidean distance between atoms directly as a cost function for the OT map. They then calculated the vector field along a straight line, moving the atomic coordinates directly to the real atomic coordinates. However, this straight-line approach may pose issues in 3D molecular conformation prediction due to the influence of spatial symmetry (see details in Appendix A), which may prevent accurate approximation of the OT mapping in 3D molecular conformation prediction.

    To address this issue, we start by considering $x_0$ as a point cloud sampled from a Gaussian distribution representing molecular coordinates, and $x_1$ as the point cloud of true molecular coordinates. We first apply the Zero CoM operation to both $x_0$ and $x_1$ to ensure translational invariance. We then use the Kabsch algorithm \cite{kabsch1976solution} to find the optimal rotation matrix that aligns $x_0$ and $x_1$ ensuring rotational invariance. Instead of using the Euclidean squared distance, we calculate the Root Mean Square Deviation (RMSD) matrix between the aligned molecules as the cost function, the RMSD for each pair of conformation $C(x_{0_i}, x_{1_j})$ is calculated using Equation \ref{eq:cost}, where $K$ is the number of atoms in the molecule, and $x_1$ is rotated using the Kabsch algorithm to achive best alignment with $x_0$.

\begin{equation}
\label{eq:cost}
C(x_{0_i}, x_{1_j}) = \sqrt{\frac{1}{K} \sum_{k=1}^{K} \left\| x_{0_{i_k}} - x_{1_{j_k}} \right\|^2}
% C(x_0, x_1) = \frac{1}{N} \sum_{i=1}^{N} \left\| R \cdot (x_{0,i} - \text{CoM}(x_0)) - (x_{1,i} - \text{CoM}(x_1)) \right\|^2
\end{equation}
% \begin{equation}
% \label{eq:cost}
% \begin{split}
% C(x_0, x_1) = \frac{1}{N} \sum_{i,j} 
% \left\| R \cdot (x_{0_i} - \text{CoM}(x_{0_i})) \right. \\
% \left. - (x_{1_j} - \text{CoM}(x_{1_j}) \right\|^2
% \end{split}
% \end{equation}

% \begin{equation}
% \label{eq:cost}
% C(x_{0_i}, x_{1_j}) = \operatorname{RMSD}(x_{0_i}, x_{1_j})
% \end{equation}

We perform OT between different conformations of each molecule. Our Equivariant OT-CFM process is as follows:
\begin{enumerate}
    \item Initialize the conformation coordinates $ x_0 $ to follow a Gaussian distribution based on the conformation coordinates $ x_1 $ from the true target distribution $ p_1 $.
    \item Calculate the cost matrix using Equation \ref{eq:cost} and then obtain the OT map matrix $ M $ with the minimal cost.
    \begin{equation}
    M = \arg\min \sum_{i,j} C(x_{0_i}, x_{1_j})
    \end{equation}
    \item Obtain the molecular coordinate pairs $(x_{0_i},x_{1_j})$ that satisfies the OT map matrix $M$.
    \item Compute the conditional probability path $ p_{t}(x|z) $ and conditional vector field $ u_{t}(x|z) $ (see Appendix B) based on the paired molecular coordinates $ (x_{0_i},x_{1_j}) $.
\end{enumerate}

\subsection{Modified Equiformer}

\subsubsection{Equiformer.}
Equiformer \cite{liao2023equiformer} is a deep learning model that combines the attention mechanism of Transformers with the SE(3) equivariance mechanism of Graph Neural Networks (GNNs). To ensure equivariance, Equiformer substitutes scalar node features with equivariant irreducible representations (irreps) features, performing equivariant operations on these irreps features, and adding equivariant graph attention to message passing. These operations include the depth-wise tensor product (DTP). and equivariant linear operations, equivariant layer normalization \cite{batzner2022}, and gate activation \cite{weiler20183d}. Furthermore, Equiformer uses nonlinear functions in attention weights and message passing, giving it stronger attention expression capabilities than typical Transformers.

\subsubsection{EquiformerV2.}
Building upon Equiformer, EquiformerV2 \cite{liao2024equiformerv2} replaces SO(3) convolutions with eSCN convolutions, enabling more efficient incorporation of higher-degree tensors. The paper introduces three architectural improvements:
\begin{enumerate}
    \item Attention Re-normalization: By introducing additional layer normalization, it stabilizes the training process when extending to higher degrees.
    \item Separable $S^2$ Activation: Replaces Gate Activation with two-layer MLP, making the training process more stable and enhancing the model's expressive capability.
    \item Separable Layer Normalization: Normalizes different degrees separately, improving the model's performance in higher degrees.
\end{enumerate}

These improvements make EquiformerV2 more efficient in handling complex three-dimensional atomic systems.

\subsubsection{Modified Equiformer.}
\begin{figure}[h]
    \centering
    \includegraphics[width=1.0\linewidth]{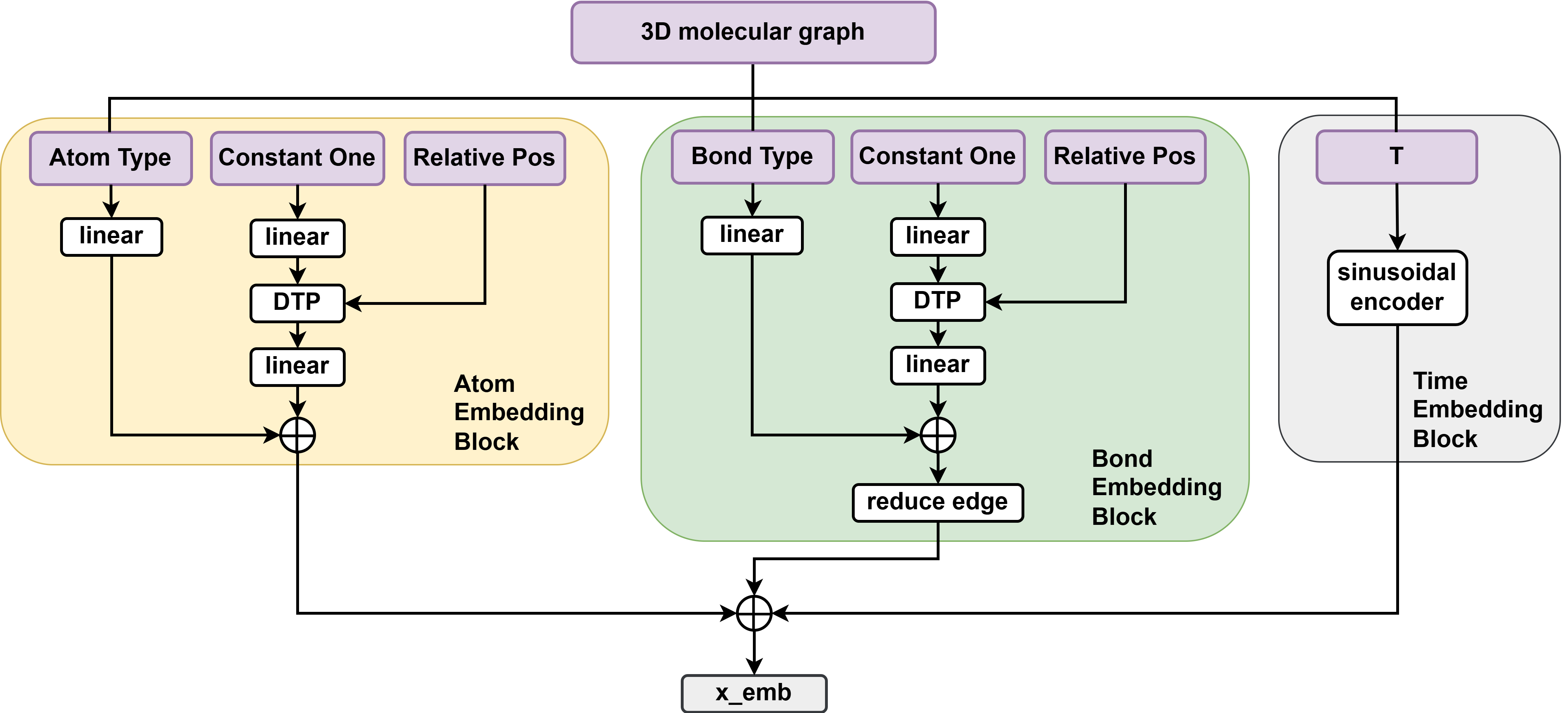}
    \caption{Atom, Bond, and Time Embedding Blocks in Modified Equiformer. We embed input 3D molecular graph with Atom Type, Bond Type, Relative Pos, and Time embeddings before transformer blocks, consisting of SO(2) equivariant graph attention and feed forward networks.}% In this figure, $\oplus$ denotes addition.
    \label{fig:embedding}
\end{figure}
EquiFlow made three modifications to EquiformerV2, adapting it for the FM training framework and tasks like single-conformation prediction and multi-conformation generation. These modifications include:
\begin{enumerate}
    \item Time Embedding: Adding time embedding $T(t)$ to all degrees and orders of irreps features $f$, enabling the model to capture temporal changes in the FM training framework.
        \begin{equation}
            f^{(L)}_{m,i} = f^{(L)}_{m,i} + T(t)_{i}
        \end{equation}
    \item Edge Feature Calculation: Changing the calculation of edge features from using Gaussian Radial Basis Function (RBF) for adjacent atom distances to using chemical bonds within molecules, allowing the model to better understand the role of chemical bonds.
    \item Prediction Head: Modifying the prediction head from aggregated 0-degree feature encoding to non-aggregated 1-degree feature encoding, since atomic 3D coordinates belong to 1-degree features.
\end{enumerate}

Figure \ref{fig:embedding} shows the embedding $x_{emb}$ used in the modified Equiformer. The embedding is a sum of atom embedding, bond embedding and time embedding. We found that using the atom embedding method to embed bond type features produces a better result than the original equiformer, where bond types are not considered. The modified Equiformer shows improved performance in molecular conformation prediction tasks within the FM training framework.

\begin{algorithm}
\caption{Training Procedure}
\label{alg:train}
\textbf{Require}: Let $v_\theta$ be the modified Equiformer. Let $x$ be a molecule, with $x_c$ denoting the coordinates of its $m$ conformations $\{x_{c_0}, x_{c_1}, \dots, x_{c_m}\}$. Here, $x_\alpha$ and $x_\beta$ represent the atom types and bond types, respectively.
\begin{algorithmic}[1]
\WHILE{Training}
    \STATE $n \leftarrow \min(m, N)$
    \STATE $x_1 \subseteq_{\text{R}} x_c \quad \text{where} \quad x_1 = \{x_{1_1}, x_{1_2}, ..., x_{1_{n}}\}$
    \STATE $x_0 \sim \mathcal{N}(0, 1)$
    \STATE $t \sim \mathcal{U}(0, 1)$
    \STATE $x_{0_i}, x_{1_j} \leftarrow \text{OT-CFM}(x_0, x_1)$
    \STATE $x_{0_i}, x_{1_j} \leftarrow (x_{0_i} - \text{CoM}(x_{0_i})), (x_{1_j} - \text{CoM}(x_{1_j}))$
    \STATE $u_t \leftarrow x_{1_j} - x_{0_i}$
    \STATE $x_t \leftarrow x_{0_i} + t \cdot u_t$
    \STATE $v_t \leftarrow v_\theta(x_t, x_\alpha, x_\beta, t)$
    \STATE $\mathcal{L}_{\text{CFM}}(\theta) \leftarrow \left\|v_t - u_t\right\|^2$
    \STATE $\theta \leftarrow \text{Update}(\theta, \nabla_\theta \mathcal{L}_{\text{CFM}}(\theta))$
\ENDWHILE
\end{algorithmic}
\end{algorithm}

\begin{algorithm}
\caption{Sampling Procedure}
\label{alg:sample}
\textbf{Require}: Let $v_\theta$ be the modified Equiformer. Let $x$ be a molecule, with $x_\alpha$ and $x_\beta$ representing the atom types and bond types, respectively. 
\begin{algorithmic}[1]
    \STATE $x_0 \sim \mathcal{N}(0, 1)$
    \STATE $x_0 \leftarrow x_0 - \text{CoM}(x_0)$
    \STATE $x_1 \leftarrow ODESolve(v_\theta, (x_0, x_\alpha, x_\beta, t), (0, 1))$
\end{algorithmic}
\end{algorithm}

\begin{table*}[t]
\centering
\begin{tabular}{lrrrrrrrr}
\toprule
Methods & \multicolumn{2}{c}{COV-R (\%) $\uparrow$} & \multicolumn{2}{c}{MAT-R (Å) $\downarrow$} & \multicolumn{2}{c}{COV-P (\%) $\uparrow$} & \multicolumn{2}{c}{MAT-P (Å) $\downarrow$} \\
& mean & median & mean & median & mean & median & mean & median \\
\midrule
RDKit & 85.1 & \textbf{100.0} & 0.235 & 0.199 & 86.8 & \textbf{100.0} & 0.232 & 0.205 \\
OMEGA & 85.5 & \textbf{100.0} & 0.177 & 0.126 & 82.9 & \textbf{100.0} & 0.224 & 0.186 \\
GeoMol & 91.5 & \textbf{100.0} & 0.225 & 0.193 & 86.7 & \textbf{100.0} & 0.270 & 0.241 \\
GeoDiff & 76.5 & \textbf{100.0} & 0.297 & 0.229 & 50.0 & 33.5 & 0.524 & 0.510 \\
Torsional diffusion & 92.8 & \textbf{100.0} & 0.178 & 0.147 & \textbf{92.7} & \textbf{100.0} & 0.221 & 0.195 \\
\midrule
EquiFlow & \textbf{95.9} & \textbf{100.0} & \textbf{0.130} & \textbf{0.079} & 91.8 & \textbf{100.0} & \textbf{0.164} & \textbf{0.108} \\
\bottomrule
\end{tabular}
\caption{Results on the GEOM-QM9 test dataset. EquiFlow outperforms all other methods in several key metrics on the GEOM-QM9 test dataset. It achieves the highest COV-R and the lowest MAT-R, indicating a superior ability to cover diverse conformations. The method also excels in precision-related metrics, with a competitive COV-P and the lowest MAT-P, highlighting its effectiveness in generating highly precise conformations.}
\label{tab:geom_qm9}
\end{table*}

\subsection{Training Procedure}
In Algorithm \ref{alg:train}, each sample $x$ in the training dataset represents a molecule containing $m$ conformations, where $m$ may vary for each molecule. The Cartesian coordinates for the conformations of each molecule are denoted by ${x_c = (x_{c_0}, x_{c_1}, ..., x_{c_m})}$. To ensure balanced training across molecules, we randomly select $n$ conformations from $x_c$ to create $x_1$, where
\begin{equation}
n=
\begin{cases}
  N & \text{if } m > N \\
  m & \text{else}
\end{cases} \quad \text{and} \quad x_1 = (x_{1_1}, x_{1_2}, ..., x_{1_{n}})
\end{equation}
here, $N$ represents the maximum number of conformations selected for training. 

A set of $n$ initial noise coordinates ${x_0 = (x_{0_1,} x_{0_2}, ..., x_{0_{n}})}$ is sampled from a Gaussian distribution $\mathcal{N}(0,1)$. Additionally, a set $t=(t_1,t_2,...,t_{n})$ is sampled from a uniform distribution $\mathcal{U}(0,1)$.
$x_0$ and $x_1$ are then passed to OT-CFM to create $(x_{0_i}, x_{1_j})$ pairs according to the cost function $C(x_0, x_1)$. then combined with $t$ to create $u_t$ and $x_t$ such that
\begin{equation}
u_t = x_{1_j} - x_{0_i} \quad \text{and} \quad x_t = x_{0_i} + tu_t
\end{equation}

Finally, the modified Equiformer $v_{\theta}$ takes $x_t$ along with the atom types and the bond types as input, then outputs $v_t$ to fit $u_t$ using MES loss function. For detailed Hyper-parameter settings related to training, please refer to Appendix C.

\subsection{Sampling Procedure}
Algorithm \ref{alg:sample} describes the sampling procedure. First, sample $x_0$ from a Gaussian distribution $\mathcal{N}(0, 1)$. This Gaussian sample is then adjusted using a Zero CoM operation to ensure it has a center of mass at the origin. The modified sample, along with the molecule's atom types $x_\alpha$, bond types $x_\beta$, and a time interval $t$, is passed to an ODE solver, which integrates the modified Equiformer model $v_\theta$ over the time interval ${[0, 1]}$ to produce the final sample $x_1$.

\begin{figure*}[h]
    \centering
    \includegraphics[width=\textwidth]{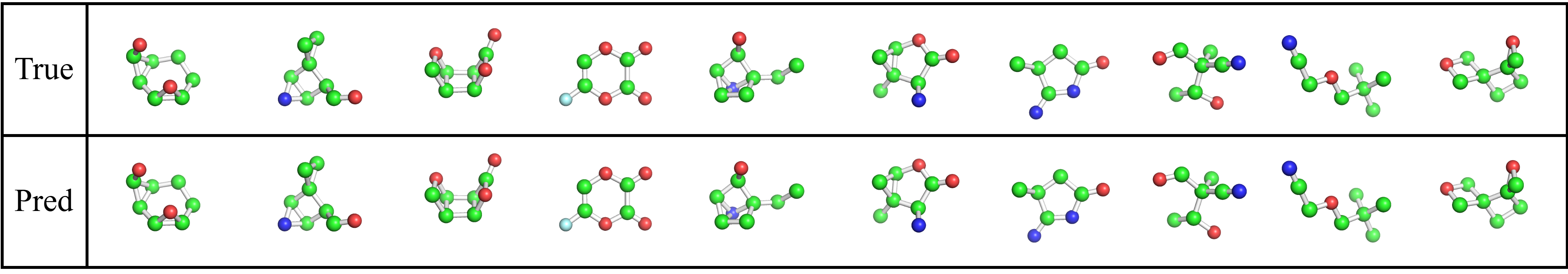}
    \caption{Samples from the single-conformation prediction on QM9 dataset. We selected 10 molecules, with SMILES from left to right as follows: C1CC2(CO2)C12COC2, CC(C)(C)COCC\#N, CC(O)C(C)(C\#N)C=O, CC1CC(=O)NC1N, CC12CC1OC(=O)C2N, CCC12CC3C(C1O)N32, O=C1OC=C(F)OC1=O, O=C1OC2C3COC3C12, O=CC1C2NC2C12CC2, OC1C2CCC3OC3C12.}
    \label{fig:samples_single}
\end{figure*}

\begin{figure*}[h]
    \centering
    \includegraphics[width=\textwidth]{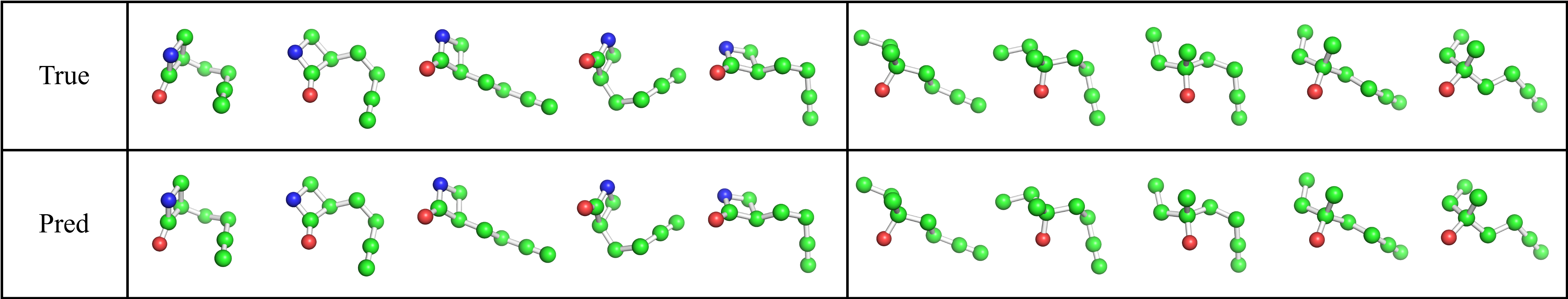}
    \caption{Samples from the multi-conformation prediction on GEOM-QM9 dataset. We selected 5 conformations from 2 molecules for display. The molecule on the left, C\#CCC[C@@H]1CNC1=O, consists of 9 conformations in total, while the molecule on the right, C\#CCCC@(O)CC, has 104 conformations.}
    \label{fig:samples_multi}
\end{figure*}

\section{Experiments}

\subsection{Experimental Setups}
\subsubsection{Evaluation Tasks.}
In this section, we evaluate EquiFlow on single-conformation and multi-conformation prediction tasks for small molecules, following the evaluation setup from previous works on 3D molecular conformation \cite{xu2022geodiff, jing2022torsional, song2024equivariant}. The single-conformation prediction task evaluates the accuracy of 3D conformation predictions on the QM9 dataset. The multi-conformation prediction task comprehensively assesses the accuracy and diversity of 3D conformation predictions on the GEOM-QM9 dataset.

\subsubsection{Datasets.}
For the single-conformation prediction task, EquiFlow employs the QM9 dataset \cite{ramakrishnan2014quantum}, which comprises 130,831 molecules, each containing up to nine heavy atoms. We remove all hydrogens and split the dataset, resulting in a training set with 100,000 molecules, a validation set with 17,748 molecules, and a test set with 13,083 molecules. For the multi-conformation prediction task, EquiFlow utilizes the GEOM-QM9 \cite{ramakrishnan2014quantum} dataset. To facilitate OT between different conformations of the same molecule, we aggregate the dataset by grouping conformations with identical SMILES notations and remove invalid conformations—those with the same SMILES notation but differing in the number or type of chemical bonds. Following the dataset split of Torsional Diffusion \cite{jing2022torsional}, the final dataset consists of a training set with 106,586 molecules comprising 1,447,427 conformations, a validation set with 13,323 molecules comprising 193,998 conformations, and a test set with 1,000 molecules comprising 13,730 conformations.
\subsubsection{Baselines.}
For the multi-conformation prediction task, we compare our results with those of RDKit, OMEGA, GeoMol, GeoDiff, and Torsional Diffusion, which are directly taken from \cite{jing2022torsional}.
\subsubsection{Evaluation Metrics.}
For the single-conformation prediction task, we measure the accuracy of the predicted molecular conformations by calculating the RMSD between the predicted molecular coordinates $x_p$ and the true molecular coordinates $x_t$. 
For the multi-conformation prediction task, we follow Geodiff \cite{xu2022geodiff} to calculate the mean and median of four RMSD-based evaluation metrics to assess the diversity and accuracy of the predicted molecular conformations. Let $S_p$ and $S_t$ represent the sets of predicted and true conformations, respectively. The coverage and matching metrics, following the traditional recall and precision measures, can be defined as:
\begin{equation}
\begin{split}
\operatorname{COV-\mathrm{R}}\left(S_{p}, S_{t}\right) &= \frac{1}{\left|S_{t}\right|} \left|\left\{x_t \in S_{t} \mid \operatorname{RMSD}(x_t, x_p) \leq \delta,\right.\right. \\
&\quad \left.\left. x_p \in S_{p}\right\}\right|
\end{split}
\end{equation}
\begin{equation}
\operatorname{MAT-\mathrm{R}}\left(S_{p}, S_{t}\right)=\frac{1}{\left|S_{t}\right|} \sum_{x_t \in S_{t}} \min _{x_p \in S_{p}} \operatorname{RMSD}(x_t, x_p)
\end{equation}
\begin{equation}
\begin{split}
\operatorname{COV-\mathrm{P}}\left(S_{p}, S_{t}\right) &= \frac{1}{\left|S_{p}\right|} \left|\left\{x_t \in S_{p} \mid \operatorname{RMSD}(x_t, x_p) \leq \delta, \right.\right. \\
&\quad \left.\left. x_p \in S_{t}\right\}\right|
\end{split}
\end{equation}
\begin{equation}
\operatorname{MAT-\mathrm{P}}\left(S_{p}, S_{t}\right)=\frac{1}{\left|S_{p}\right|} \sum_{x_t \in S_{p}} \min _{x_p \in S_{t}} \operatorname{RMSD}(x_t, x_p)
\end{equation}
where $\delta$ is a pre-defined threshold, $\delta$ is set as 0.5Å for the GEOM-QM9 dataset. The size of $ S_p $ for each molecule is set to twice the size of $ S_t $. The COV score reflects the percentage of predicted conformations that are deemed successful, where a successful prediction is defined as having an RMSD below the threshold $\delta$. The MAT score calculates the average RMSD between conformations in one set and their nearest counterparts in another. Typically, recall measures diversity, while precision measures accuracy. Higher COV rates or lower MAT scores suggest the prediction is more realistic.

\subsection{Experimental Results}

EquiFlow performed multi-conformation prediction based on the GEOM-QM9 dataset. The prediction results on the test dataset ($\delta$ = 0.5Å) is shown in Table \ref{tab:geom_qm9}. These results suggest that EquiFlow significantly improves both the diversity and accuracy of molecular conformation predictions compared to other state-of-the-art methods. Apart from multi-conformation prediction, EquiFlow also achieved an excellent RMSD of 0.17Å on the QM9 test set for single conformations, demonstrating it's outstanding ability on both single-conformation and multi-conformation prediction tasks.

\subsection{Conformation Visualization}
To better highlight the performance of EquiFlow, we visualized the predicted molecular conformations for both tasks to provide a qualitative comparison (Figures \ref{fig:samples_single}, \ref{fig:samples_multi}). EquiFlow successfully captures the complex distribution of molecular conformations, achieving accurate predictions while also maintaining state-of-the-art diversity.

\section{Conclusion}
We propose EquiFlow, a novel framework for predicting 3D molecular conformations with a focus on accuracy and diversity. The model leverages a combination of Equivariant OT-CFM and a modified Equiformer to handle atomic-level features and 3D spatial symmetry. The incorporation of OT provides an unique approach for mapping molecular conformations, while modifications to the Equiformer architecture enhance performance in tasks involving both single-conformation and multi-conformation prediction. Our experimental results on the QM9 and GEOM-QM9 datasets demonstrated that EquiFlow achieves state-of-the-art results compared to previous methods, ensuring accurate predictions while maintaining excellent diversity. In future work, we plan to further optimize the model by incorporating the latest advances in CFM methods and extend our approach to conformation prediction for larger molecules.

% \newpage
\bibliography{EquiFlow}

\appendix
\include{append}

\end{document}

%% file: append.tex
% \section*{Appendix}

\section{A. 3D Molecular Symmetry Problem}
\subsection{SE(3)-Equivariant and Irreducible Representations(irreps)}
For any input \( x \in X \), output \( y \in Y \), and group element \( g \in G \), if the condition \({f(D_X(g)x) = D_Y(g)f(x)}\) is satisfied, then we call the function \( f \) equivariant. Here, \( X \) and \( Y \) are vector spaces, \( f \) is a mapping function between them, and \( D_X(g) \) and \( D_Y(g) \) are the transformation matrices parameterized by \( g \) in \( X \) and \( Y \). In the context of 3D atomic graphs, we specifically focus on the special Euclidean group \( SE(3) \), which arises from 3D translations and rotations. In this case, the input features and the learnable functions possess \( SE(3) \)-Equivariant properties. Specifically, the transformations \( D_X(g) \) and \( D_Y(g) \) can be represented by a translation \( t \) and an orthogonal matrix \( R \) that denotes a rotation or reflection. When these transformations are applied to the input, if the output undergoes the corresponding equivalent transformation, satisfying \( Rf(x) = f(Rx) \), then we call the function \( f \) equivariant with respect to the rotation or reflection \( R \).

For the three-dimensional Euclidean group \( SE(3) \), under translation and rotation transformations, vectors change with rotation while scalars remain invariant. To address translational symmetry, we follow \cite{xu2022geodiff} and use the Zero Center-of-Mass(Zero CoM) Operation. For rotational equivariance, it is necessary to introduce the irreducible representations of \( SO(3) \), which are the irreducible representations of rotations.

In a given vector space, any representation of the \( SO(3) \) group can be decomposed into a series of minimal, indivisible transformation matrices, called irreducible representations (irreps). Specifically, for a group element \( g \in SO(3) \), there exists an irreducible matrix \( D_L(g) \) of dimension \( (2L+1) \times (2L+1) \) (also known as the Wigner-D matrix), which acts on a \( (2L+1) \)-dimensional vector space. These irreducible representations are indexed by non-negative integers \( L = 0, 1, 2, \ldots \), with the \( L \)-th degree irreducible representation having a dimension of \( (2L+1) \). Here, \( L \) can be interpreted as an angular frequency, which determines the rate of change of the vector in the rotating coordinate system. The \( D_L(g) \) for different \( L \) acts on independent vector spaces. The vectors transformed by \( D_L(g) \) are called \( L \)-degree vectors. Specifically, \( L = 0 \) (dimension 1) corresponds to a scalar; \( L = 1 \) (dimension 3) corresponds to a vector; \( L = 2 \) (dimension 5) corresponds to a second-degree vector. The elements of an \( L \)-degree vector \( f^{(L)} \) are indexed by the integer \( m \), denoted as \( f_m^{(L)} \) (with \( m \) ranging from \( -L \le m \le L \)).

If multiple \( L \)-degree vectors are concatenated, they form irreducible features with \( SE(3) \)-equivariance. Generally, these irreducible features can be represented as \( f_{c,m}^{(L)} \), where \( 0 \le L \le L_{\text{max}} \), \( 0 \le c \le C_L - 1 \), and \( C_L \) denotes the number of channels for the \( L \)-degree vector. This means that in each \( L \)-degree vector, there are \( C_L \) different vectors, each containing \( m \) elements. Different channels of the \( L \)-degree vectors are parameterized by different weights but transformed using the same \( D_L(g) \). Significantly, conventional scalar features correspond to 0-degree vectors.

\subsection{Spherical Harmonics and Tensor Product}
Spherical Harmonics (SH) achieve the mapping from the unit sphere to the irreducible representation $D^l$, which can be expressed by the function $Y^{(L)}$. Using spherical harmonics, a vector $\vec{r}$ in the Euclidean space $\mathbb{R}^3$ can be mapped to an $L$-degree vector $f^{(L)}$, i.e., $f^{(L)} = Y^{(L)} \left(\frac{\vec{r}}{\|\vec{r}\|}\right)$. SE(3)-Equivariant networks update irreducible features by passing messages of transformed irreducible features between nodes. To interact different L-type vectors during the message-passing process, we use the Tensor Product, which extends multiplication to equivariant invariant features \cite{liao2023equiformer}. In $SO(3)$, the tensor product combines an $L_1$-degree vector $f^{(L_1)}$ and an $L_2$-degree vector $g^{(L_2)}$ using CG (Clebsch-Gordan) coefficients to generate an $L_3$-degree vector $h^{(L_3)}$:
\begin{equation}
h_{m_3}^{(L_3)} = (f^{(L_1)} \otimes g^{(L_2)})_{m_3}
\end{equation}
% \begin{equation}
% (f^{(L_1)} \otimes g^{(L_2)})_{m_3} = \sum_{m_1 = -L_1}^{L_1} \sum_{m_2 = -L_2}^{L_2} C_{(L_1, m_1)(L_2, m_2)}^{(L_3, m_3)} f_{m_1}^{(L_1)} g_{m_2}^{(L_2)}
% \end{equation}
\begin{equation}
\begin{split}
(f^{(L_1)} \otimes g^{(L_2)})_{m_3} &= \sum_{m_1 = -L_1}^{L_1} \sum_{m_2 = -L_2}^{L_2} C_{(L_1, m_1)(L_2, m_2)}^{(L_3, m_3)} \\
&\quad \times f_{m_1}^{(L_1)} g_{m_2}^{(L_2)}
\end{split}
\end{equation}
where $m_1$, $m_2$, and $m_3$ represent the order, corresponding to the $m_1$-th, $m_2$-th, and $m_3$-th elements of $f^{(L_1)}$, $g^{(L_2)}$, and $h^{(L_3)}$, respectively. The CG coefficients are non-zero only when $|L_1 - L_2| \le L_3 \le |L_1 + L_2|$, which constrains the type of the output vector. Thus, the type of $h^{(L_3)}$ depends on the types of $f^{(L_1)}$ and $g^{(L_2)}$. For example, when $f^{(L_1)}$ is a scalar (i.e., $L_1 = 0$) and $g^{(L_2)}$ is a vector ($L_2 = 1$), $h^{(L_3)}$ is a vector ($L_3 = 1$) formed by the product of $f^{(L_1)}$ and $g^{(L_2)}$.

\section{B. Conditional Flow Matching with Optimal Transport}
Where the conditional vector field $ u_{t}(x|z) $ and conditional probability path $ p_{t}(x|z) $ yield the vector field $ u_{t}(x) $ and the corresponding probability path $ p_{t}(x) $:
\begin{equation}
p_{t}(x) = \int p_{t}(x|z) p(z) \, dz \quad 
\end{equation}
\begin{equation}
\quad u_{t}(x) = \int \frac{p_{t}(x|z) u_{t}(x|z)}{p_{t}(x)} p(z) \, dz
\end{equation}
Here, $p(z)$ is an arbitrary conditional distribution independent of $x$ and $t$. For detailed derivations, refer to \cite{lipman2023flow, tong2024improving}. Following the methodology in \cite{tong2024improving}, $ u_{t}(x|z) $ and $ p_{t}(x|z) $ are parameterized as:
\begin{equation}
z = (x_{0}, x_{1}) \quad \text{and} \quad p(z) = \pi(x_{0}, x_{1})
\end{equation}
\begin{equation}
u_{t}(x|z) = x_{1} - x_{0}
\end{equation}
\begin{equation}
\quad p_{t}(x|z) = \mathcal{N}\left(x \mid t \cdot x_{1} + (1-t) \cdot x_{0}, \sigma^{2}\right)
\end{equation}

\section{C. Experimental Details}
In this section, we introduce the details of our experiments. We first obtain the feature information of the 3D molecular graph, including atom types, bond types, and coordinates $x_1$. Next, we generate a Gaussian initial coordinate $x_0$. FM is then performed between $x_0$ and $x_1$ to obtain the time-dependent probability path $x_T$ and the corresponding vector field $u_T$. Finally, we design a Modified Equiformer network to regress this vector field. All hyper-parameters in training of EquiFlow are summarized in Table 2.

\begin{table}[t]
\centering
\begin{tabular}{ll}
\toprule
Hyper-parameters & Value \\
\midrule
Number of Transformer blocks & 6 \\
Maximum degree & [6] \\
Sigma of cfm & 0.0 \\
Remove hydrogen & true \\
Optimizer & AdamW \\
\midrule
\multicolumn{2}{c}{Single-conformation} \\
Flow matcher & cfm \\
Train batch size & 256 \\
Eval batch size & 256 \\
Learning rate & 0.0005 \\
Weight decay & 0.005 \\
\midrule
\multicolumn{2}{c}{Multi-conformation} \\
Sample coefficient & 2 \\
Flow matcher & otcfm \\
Train batch size & 128 \\
Eval batch size & 32 \\
Maximum number of conformations & 20 \\
RMSD threshold for coverage & 0.5 \\
Learning rate & 0.001 \\
Weight decay & 0.005 \\
\bottomrule
\end{tabular}
\caption{Hyper-parameters for the single-conformation and multi-conformation prediction tasks.}
\label{tab:single}
\end{table}